\documentclass[letterpaper, 10pt, conference]{ieeeconf}
\IEEEoverridecommandlockouts  

\usepackage{xurl} 
\usepackage{float}
\usepackage{comment}
\usepackage[hidelinks]{hyperref}
\usepackage{cite}
\usepackage{amsmath,amssymb,amsfonts}
\usepackage{algorithmic}
\usepackage{graphicx}
\usepackage{textcomp}
\usepackage{xcolor}
\def\BibTeX{{\rm B\kern-.05em{\sc i\kern-.025em b}\kern-.08em
    T\kern-.1667em\lower.7ex\hbox{E}\kern-.125emX}}

\usepackage{tikz}
\usetikzlibrary{patterns,decorations.pathmorphing,positioning,arrows.meta, positioning,calc,shapes.geometric}
\usepackage{stfloats} 
\begin{document}

\title{A Kalman Filter-Based Disturbance Observer for Steer-by-Wire Systems}

\author{Nikolai Beving$^{1}$, Jonas Marxen$^{2}$, Steffen Müller$^{2}$, Johannes Betz$^{1}$
\thanks{$^{1}$ All authors are with the Professorship of Autonomous Vehicle Systems, Technical University of Munich, 85748 Garching, Germany; Munich Institute of Robotics and Machine Intelligence (MIRMI). Contact: \{marc.kaufeld, mattia.piccinini, johannes.betz\}@tum.de}
\thanks{$^{2}$ J. Marxen, S. Müller are with the Chair of Automotive Engineering, Institute of Land and Sea Transport, Technical University of Berlin, 13355 Berlin, Germany.}}

\maketitle

\begin{abstract}
Steer-by-Wire systems replace mechanical linkages, which provide benefits like weight reduction, design flexibility, and compatibility with autonomous driving. However, they are susceptible to high-frequency disturbances from unintentional driver torque — known as driver impedance — which can degrade steering performance. Existing approaches either rely on direct torque sensors — which are costly and impractical — or lack the temporal resolution to capture rapid, high-frequency driver-induced disturbances. We address this limitation by designing a Kalman filter-based disturbance observer that estimates high-frequency driver torque using only motor state measurements. We model the driver’s passive torque as an extended state using a PT1-lag approximation and integrate it into both linear and nonlinear Steer-by-Wire system models. In this paper, we present the design, implementation and simulation of this disturbance observer with an evaluation of different Kalman filter variants.
Our findings indicate that the proposed disturbance observer accurately reconstructs driver-induced disturbances with only minimal delay ($\sim 14~\mathrm{ms}$). We show that a nonlinear extended Kalman Filter outperforms its linear counterpart in handling frictional nonlinearities, improving estimation during transitions from static to dynamic friction. Given the study's methodology, it was unavoidable to rely on simulation-based validation rather than real-world experimentation. Further studies are needed to investigate the robustness of the observers under real-world driving conditions.

\end{abstract}


\section{Introduction}
\begin{figure}[t]
\centering
\includegraphics[width=0.89\columnwidth]
{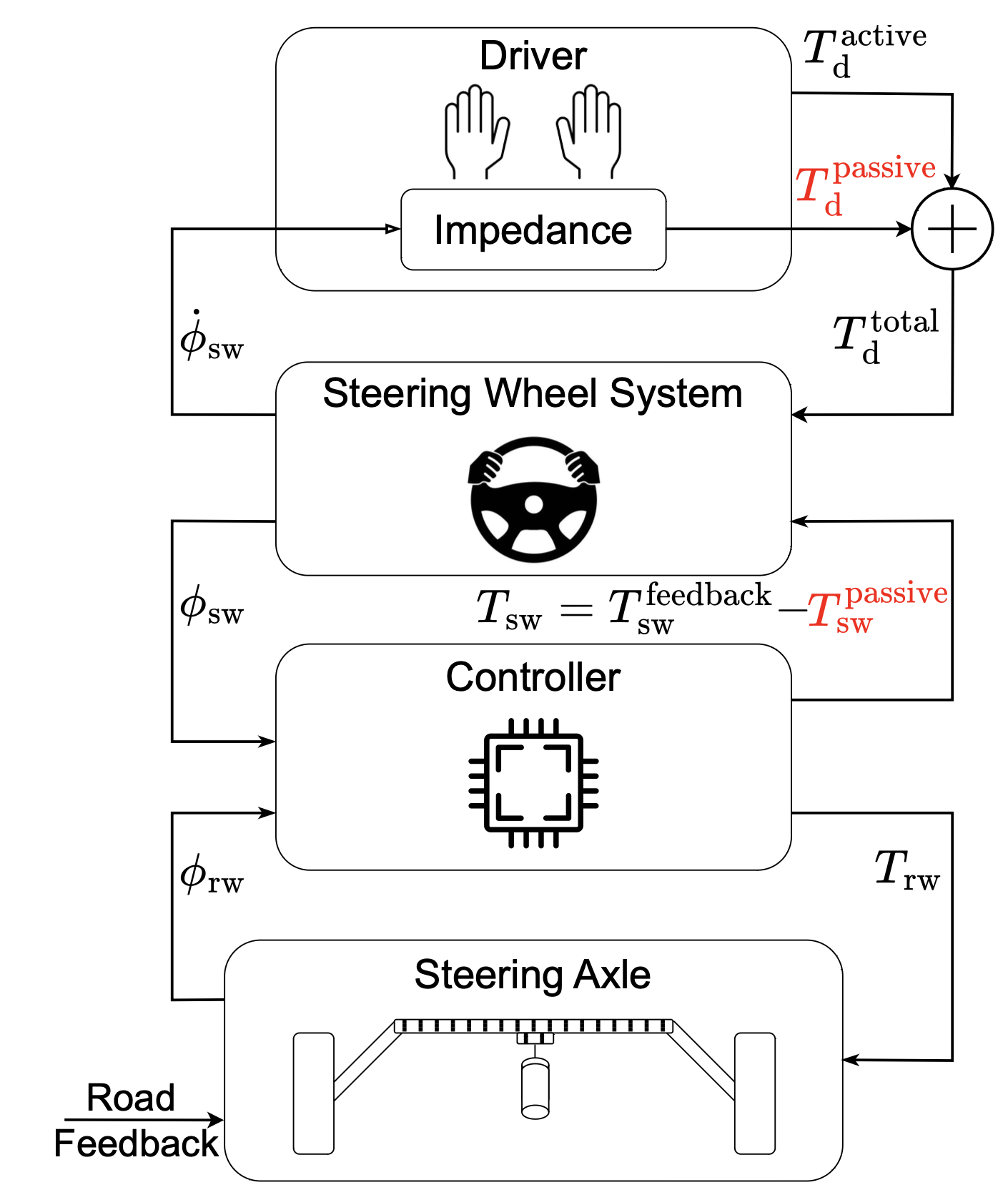}
\caption{Schematic architecture of a SbW-system with consideration of driver impedance and resulting passive driver torque rejection.}
\vspace{0.0cm}
\label{fig:sbw_konzept_v4.drawio.pdf}
\end{figure}

Steer-by-Wire (SbW) systems benefit from a purely electrical connection between the steering wheel and the vehicle front wheels. These systems consist of two main subsystems: the hand-wheel (HW) module and the road-wheel (RW) module, each comprising an electric motor and a gear, along with other components. The HW module is responsible for transmitting driving feedback torque to the driver while simultaneously sensing the driver's steering commands. Similarly, the RW module receives information from the HW module, steers the tires accordingly, and senses the forces and movements of the wheels, transmitting this information back to the HW module as schematically depicted in \mbox{Fig. \ref{fig:sbw_konzept_v4.drawio.pdf}} with $\phi$ denoting the steering and road wheel angles, and torque $T$ referring to the steering wheel, road wheels, and the driver. \cite{mortazavizadeh_recent_2020}

SbW systems require two coupled control loops: one for the HW actuator and one for the RW actuator. The resulting flexibility in design, the complexity of control approaches, and safety-related challenges have so far hindered SbW systems from becoming established in the automotive industry. Despite these challenges, SbW systems offer significant advantages, including enhanced crash safety, reduced weight and space requirements, and improved design flexibility for hardware components. Additionally, SbW enables advanced driving functions, such as customizable steering feedback, variable steering ratios, and compatibility with autonomous driving requirements \cite{mortazavizadeh_recent_2020}\cite{ewald_regelung_2022}.
The main contribution of this paper is a novel disturbance observer (DOB) that
\begin{itemize}
    \item detects high-frequency disturbances and enables the controller to reject them, as shown in Fig. \ref{fig:sbw_konzept_v4.drawio.pdf},
    \item uses a Kalman-based approach for both linear and nonlinear dynamics,
    \item assumes that only the motor states — angle, angular velocity, and torque — are available, reflecting practical constraints in the automotive industry,
    \item can be applied to both HW and RW module torque estimation,
    \item has its performance evaluated in both linear and nonlinear approaches through simulation, where the effectiveness of disturbance rejection is assessed.
\end{itemize}


%
%

%
%


\section{State of the Art}

\subsection{Problem Formulation}
On both the RW and HW modules, disturbances can affect the controlled system and reduce overall performance. For the HW module, disturbances are primarily induced by the driver. These can be categorized into active, deliberate steering wheel actions, which are typically low-frequency, and passive, unintentional high-frequency steering wheel \mbox{actions \cite{ewald_regelung_2022}.} The latter occur when the driver maintains their hands on the steering wheel while high-frequency road feedback (e.g., cobblestone pavement or potholes) is transmitted to the steering wheel. In such cases, the driver unintentionally adds impedance to the HW module, potentially degrading control performance \cite{bajcinca_robust_2005}. As noted in \cite{ewald_regelung_2022}, drivers' intentional steering maneuvers are performed below 2 Hz; therefore, higher frequencies can be considered unwanted disturbances.

The identical problem can be observed with the RW module. When driving on an uneven road, varying forces can act on the wheels and disturb the system. For high-frequency forces, it is not always desirable for these forces to cause a change in the front wheel steering angle. Therefore, depending on the SbW design, rejecting such disturbances might be beneficial, especially if smooth and comfortable driving is the goal.


\subsection{Related Work}
The concept of considering the driver as an impedance and consequently separating the driver steering torque into active and passive parts has mostly been used in the context of ADAS and cooperative driving as in \cite{winner_handbuch_2024}, and has rarely been used in the context of steering control. Thus, driver steering torque observers generally focus on observing the total driver torque without aiming to capture highly dynamic disturbances.

For conventional steering systems like electric power steering, various studies on driver steering torque estimation have been conducted. Methods such as sliding mode observers \cite{marouf_driver_2010}, $H_\infty$/$H_2$ proportional-integral observers \cite{yamamoto_driver_2015}, and unknown input observers \cite{nguyen_unknown_2019} have been studied.

In the context of SbW, which comes with vastly different mechanics and limited information that can be utilized \cite{huang_robust_2020}, only few observer concepts were subject to research so far. The Luenberger-framework was applied in \cite{zhou_driver_2021,zhou_mixed_2022,wang_online_2018,huang_robust_2020} using different approaches with respect to performance criteria such as $H_2$, $H_\infty$, and $L_2$. \cite{zhou_driver_2021}, \cite{huang_robust_2020} and \cite{zhou_mixed_2022} could successfully observe the torque but did not aim for high dynamics, showing a significant time delay in the estimation. The authors of \cite{wang_online_2018} assume measurable torque, which is not a practical approach in the industry. Thus, the approaches so far are not suitable for detecting highly dynamic impedance-based disturbances without using a torque sensor. 

\section{Methodology}
\subsection{Linear Dynamics of the Hand Wheel Module}
\label{subsec:lindyn}
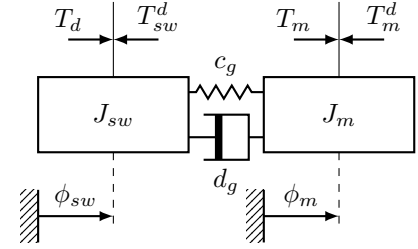
\begin{figure}[htbp]
    \centering
    \begin{tikzpicture}[every node/.style={outer sep=0pt},thick,
     mass/.style={draw,thick},
     spring/.style={thick,decorate,decoration={zigzag,pre length=0.1cm,post
     length=0.1cm,segment length=6}},
     ground/.style={fill,pattern=north east lines,draw=none,minimum
     width=0.75cm,minimum height=0.3cm},
     dampic/.pic={\fill[white] (-0.1,-0.3) rectangle (0.3,0.3);
     \draw (-0.3,0.3) -| (0.3,-0.3) -- (-0.3,-0.3);
     \draw[line width=1mm] (-0.1,-0.3) -- (-0.1,0.3);}]
 
      \node[mass,minimum width=2cm,minimum height=1cm] (m1) {$J_{sw}$};
      \node[mass,minimum width=2cm,minimum height=1cm, right=1cm of m1] (m2) {$J_{m}$};
    
      \draw[spring]  ([yshift=3mm]m1.east) coordinate(aux) -- (m2.west|-aux) node[midway,above=1mm]{$c_g$};
    
      \draw ([yshift=-3mm]m1.east) coordinate(aux')
        -- (m2.west|-aux') pic[midway]{dampic} node[midway,below=3mm]{$d_g$};
      
       \draw[thin] (m1.north) -- ++ (0,1) coordinate[midway](aux1);
       \draw[latex-] (aux1) -- ++ (-0.6,0) node[above]{$T_d$}; 
       \draw[latex-] (aux1) -- ++ (+0.6,0) node[above]{$T^d_{sw}$}; 
       \draw[thin,dashed] (m1.south) -- ++ (0,-1) coordinate[pos=0.85](aux'1);
       \draw[latex-] (aux'1) -- ++ (-1,0) node[midway,above]{$\phi_{sw}$}
        node[left,ground,minimum height=7mm,minimum width=1mm] (g'1){};
       \draw[thick] (g'1.north east) -- (g'1.south east);
      
       \draw[thin] (m2.north) -- ++ (0,1) coordinate[midway](aux2);
       \draw[latex-] (aux2) -- ++ (-0.6,0) node[above]{$T_{m}$}; 
       \draw[latex-] (aux2) -- ++ (+0.6,0) node[above]{$T^d_{m}$}; 
       \draw[thin,dashed] (m2.south) -- ++ (0,-1) coordinate[pos=0.85](aux'2);
       \draw[latex-] (aux'2) -- ++ (-1,0) node[midway,above]{$\phi_{m}$}
        node[left,ground,minimum height=7mm,minimum width=1mm] (g'2){};
       \draw[thick] (g'2.north east) -- (g'2.south east);
      
    \end{tikzpicture}
    \caption{2-mass spring damper HW system model (no gear displayed)}
    \label{fig:HWASys}
\end{figure}

The HW module is modeled as a two-mass spring-damper system as per Fig. \ref{fig:HWASys}, consisting of the electric motor and the steering wheel as masses with inertia $J_m$ and $J_{sw}$, respectively. Each mass has a linear damping component with coefficients $d_{m}$ and $d_{sw}$, representing the rotational friction torques $T^d_{m}$ and $T^d_{sw}$. Driver torque $T_d$ and motor torque $T_m$ act directly on the respective masses via the input $u = \begin{bmatrix} T_d & T_m \end{bmatrix}^T$. The gear and mechanics connecting the masses are modeled using a spring and a damping component with coefficients \( c_g \) and \( d_g \). The connection is modeled inertialess, with their inertia already accounted for in the masses $J_m$ and $J_{sw}$.
All parameters are referenced to the coordinate system of the steering wheel. 
Accordingly, the state vector $x$ comprises the angle and angular velocity for both steering wheel and motor: $\phi_{sw}$, $\dot{\phi}_{sw}$,  $\phi_m$ and $\dot{\phi}_m$. 
As $\phi_m$ and $\dot{\phi}_m$ are assumed to be measurable, the output equation results in \eqref{eq:output}.
\begin{equation}
\label{eq:2ms}
\setlength\arraycolsep{0.7pt}
\dot{x}
=
\begin{bmatrix}
0 & 1 & 0 & 0 \\
\frac{-c_{g}}{J_{sw}} & \frac{-d_{g} - d_{sw}}{J_{sw}} & \frac{c_{g}}{J_{sw}} & \frac{d_{g}}{J_{sw}} \\
0 & 0 & 0 & 1 \\
\frac{c_{g}}{J_{m}} & \frac{d_{g}}{J_{m}} & \frac{-c_{g}}{J_{m}} & \frac{-d_{g} - d_{m}}{J_{m}} \\
\end{bmatrix}
x
+
\begin{bmatrix}
0 & 0 \\
\frac{1}{J_{sw}} & 0 \\
0 & 0 \\
0 & \frac{1}{J_{m}} \\
\end{bmatrix}
u
\end{equation}

\begin{equation}
\label{eq:output}
y = 
\begin{bmatrix}
0 & 0 & 1 & 0 \\
0 & 0 & 0 & 1
\end{bmatrix}
x
\end{equation}

\subsection{Nonlinear Dynamics of the Hand Wheel Module}
\label{subsec:nonlindyn}
To obtain a more accurate model of the HW module, it is essential to consider nonlinear terms, as friction and stiffness are mostly nonlinear, especially in multi-body \mbox{mechanics \cite{zucca_nonlinear_2014}.} For the friction of motor and steering wheel, a Stribeck curve based term as in 
\eqref{eq:stribec} is used \cite{ewald_regelung_2022}. The equation models the friction torque \( T^d(\dot{\phi}) \) as a function of the angular \mbox{velocity \( \dot{\phi} \),} capturing the nonlinear behavior between static and kinetic friction. It includes the static friction torque \( T_s \), the kinetic friction torque \( T_k \), and a viscous damping term \( T_{v} \dot{\phi} \) \cite{andersson_friction_2009}. The exponential term models the transition between static and dynamic friction, with \( \dot{\phi}_c \) representing the characteristic velocity at which this transition occurs, and \( \delta \) shaping the steepness of the curve \cite{andersson_friction_2009}. Here, $s$ denotes the sign function.
\begin{equation}
\label{eq:stribec}
T^d(\dot{\phi}) = s(\dot{\phi}) 
\left( d^k + (d^s - d^k) e^{-\left|\frac{\dot{\phi}}{\dot{\phi}_c}\right|^{\delta}} \right) 
+ d^v \dot{\phi}
\end{equation}

To more accurately model the stiffness and damping of the gear, nonlinear terms are added to the previously introduced linear components $c_{g} \cdot (x_3-x_1)$ and 
$d_{g} \cdot (x_4 - x_2)$, as described by~\eqref{eq:2ms}. Since the gear comprises multiple components, each contributing distinct stiffness and friction characteristics, the system exhibits significant nonlinearity. Consequently, the nonlinear equation~\eqref{eq:nonlin_friction} can provide a more precise representation of the gear's behavior compared to a purely \mbox{linear model.} 
\begin{equation} \label{eq:nonlin_friction} 
T = k_1 \Delta \omega + k_2 |\Delta \omega|^{\alpha} s \left( \Delta \omega \right) + c_1 \Delta \dot{\omega} + c_2 |\Delta \dot{\omega}|^{\beta} s \left( \Delta \dot{\omega} \right) 
\end{equation}

Using the Stribeck curve alongside the nonlinear terms for the gear's stiffness and friction, 
the resulting torque contributions are given by \eqref{eq:damping_terms}. 
\begin{equation}
\label{eq:damping_terms}
\begin{aligned}
T_g^c &= c_{g}^{1} \cdot (x_3-x_1) + c_{g}^{2} \cdot |x_3 - x_1|^\alpha \cdot s (x_3 - x_1) \\
T_g^d &= d_{g}^{1} \cdot (x_4 - x_2) + d_{g}^{2} \cdot |x_4 - x_2|^\beta \cdot s (x_4 - x_2) \\
T_{sw}^d &= s (x_2) \big( d_{sw}^{k} + (d_{sw}^{s} - d_{sw}^{k}) e^{-\left|\frac{x_2}{\omega_{sw}^{s}}\right|^{\delta}}
\big) + d_{sw}^v x_2 \\
T_{m}^d &= s (x_4) \big(
d_{m}^{k} + (d_{m}^{s} - d_{m}^{k}) e^{-\left|\frac{x_4}{\omega_{m}^{s}}\right|^{\delta}}
\big) + d_m^v x_4 \\
\end{aligned}
\end{equation}
With the terms above, the system of nonlinear differential equations and the respective measurement function result in \eqref{eq:2ms_nonlin} and \eqref{eq:2ms_nonlinoutput}, respectively. 
\begin{align}
\label{eq:2ms_nonlin}
\begin{split}
\dot{x}_1 &= x_2 \\
\dot{x}_2 &= \frac{T_g^c + T_g^d - T_{sw}^d + u_1}{J_{sw}} \\
\dot{x}_3 &= x_4 \\
\dot{x}_4 &= \frac{-T_g^c - T_g^d - T_{m}^d + u_2}{J_m}
\end{split} \\
\label{eq:2ms_nonlinoutput}
y &=
\begin{bmatrix}
x_3 \\
x_4
\end{bmatrix}
\end{align}

\subsection{Modelling the Driver Torque}
\label{subsec:modeldriver}
Numerous investigations have been conducted in the context of modeling human behavior and its neuromuscular system to predict movements, as in \cite{valero-cuevas_computational_2009, fuchs_modeling_2023, fuchs_modeling_2022, gillam_predictive_2024, ting_review_2012}. This is a complex field of research, and techniques often require extensive resources to be implemented. Furthermore, individual human and driving characteristics, such as variations in how one holds the steering wheel, highly affect the \mbox{prediction \cite{bajcinca_robust_2005, ewald_regelung_2022}.} Therefore, this work adopts a more pragmatic approach by applying a Kalman-based estimation framework with a simplified prediction step. Nevertheless, by leveraging mechanical knowledge and measurements, this approach still provides a reliable estimation, as demonstrated later \mbox{in this paper.}

The transition from passive driving behavior to intentional low-frequency steering maneuvers is primarily driven by neuromuscular activation. Muscle activation and deactivation can both be modeled as a PT1-lag \cite{niu_model_2020}, making this approach suitable for short-term modeling of the driver torque. 
The driver's unintentional high-frequency steering wheel actions are mostly based on the additional mass of the arms adding inertia, damping and stiffness to the steering wheel. The driver's arms can simply be modeled as a two-mass spring damper system \cite{bajcinca_robust_2005, niu_model_2020, ewald_regelung_2022}. Assuming a high damping coefficient with negligible vibration, a PT1-lag element can also be used here to approximate this interaction, as its behavior roughly resembles that of a PT2-lag with a high damping coefficient. 

Additionally, when modeling the driver's behavior on high-frequency steering wheel excitations, the so-called muscle stretch reflex is relevant. The stretch reflex refers to the contraction of a muscle in response to its passive stretching \cite{bhattacharyya_stretch_2017,niu_model_2020}, resulting in a damping of the steering wheel. Again, as this depends on muscle activation, it can be approximated as PT1-like behavior \cite{niu_model_2020}. 

Under these considerations, a PT1 element provides a straightforward approximation of the driver torque \(T_d\). It captures the short-term smoothing characteristics observed in human behavior and physical dynamics, rather than serving as a genuine predictor of torque. The continuous representation of this element, shown in Equation~\eqref{eq:pt1}, describes a first-order system. Here, the time constant \(T\) governs the rate of response to changing inputs, while the gain \(K\) determines the steady-state proportionality between input and output.
\begin{equation}
\label{eq:pt1}
\dot{x} = -\frac{1}{T} \cdot x + \frac{K}{T} \cdot u
\end{equation}
The estimated driver torque \(T_d\) is fed back into the \mbox{input \(u\)} of the PT1 element. This feedback structure allows the PT1 element to serve as a bridge, enabling the Kalman filter-based DOB to estimate the input torque — something inherently not possible without such a model. To ensure proper design of the DOB, the prediction of \(T_d\) must be weighted less heavily than the other states and measurements by appropriately selecting the covariance matrices.

\subsection{Discrete Kalman Filter}
\label{discreteKF}
The Kalman filter is an optimal recursive estimator for linear dynamical systems with Gaussian noise, described in the state-space representation:
\begin{equation}
x_k = A x_{k-1} + B u_k + w, \quad y_k = C x_k + v,
\end{equation}
where \( A \) is the state transition matrix, \( B \) is the control input matrix, and \( C \) is the measurement matrix \cite{ortega_contreras_kalman-like_2021, simon_optimal_2006}. The terms \( w \) and \( v \) represent Gaussian process and measurement noise, respectively, assumed to be uncorrelated, have zero mean and covariance matrices \( Q \) and \( R \) \cite{simon_optimal_2006}. The Kalman filter minimizes the mean of the squared error to obtain approximations from a model and measurements \cite{simon_optimal_2006}. 
At each iteration, the filter computes the \textbf{a posteriori} state estimate \( \hat{x}_{k|k} \) by combining the \textbf{a priori} prediction \( \hat{x}_{k|k-1} \), based on the system dynamics, with a weighted measurement correction term $K_k (z_k - C \hat{x}_{k|k-1})$, where \( K_k \) is the Kalman gain \cite{chui_kalman_2017}. This gain minimizes the a posteriori error covariance \( P_{k|k} \) as per \eqref{eq:covariance_update}, balancing the uncertainty between the model prediction and the measurement \cite{welch_introduction_2006}. It scales the measurement residual, which is the difference between the actual measurement \( z_k \) and the predicted output \( C \hat{x}_{k|k-1} \) as per \eqref{eq:kalman_gain} \cite{welch_introduction_2006}. The second subscript index \( _{k-1} \) respectively~\( _k \) correspond to the prediction step (\text{a priori}, \( _{k-1} \)), where the state is estimated solely based on the system model, respectively the correction step (\text{a posteriori}, \( _k \)), where the measurement is incorporated. Thus, the algorithm consists of two steps \cite{welch_introduction_2006}:

\textbf{1) Prediction}
\begin{align}
\hat{x}_{k|k-1} &= A \hat{x}_{k-1|k-1} + B u_k, \label{eq:prediction_state} \\
P_{k|k-1} &= A P_{k-1|k-1} A^T + Q, \label{eq:prediction_covariance}
\end{align}
where $Q$ denotes the process noise covariance matrix, and the diagonal entries \(\sigma_i^2\) represent the model error variances for each state, assuming uncorrelated process noise \cite{walach_kalman-bucy-filter_2013}. Accurate estimation requires that \(Q\) correctly characterizes the actual process noise \(w\). As described in \eqref{eq:prediction_covariance}, the a priori error covariance \( P_{k|k-1} \) represents the uncertainty of the predicted state before incorporating the measurement. It propagates the previous covariance \( P_{k-1|k-1} \) through the system dynamics using the state transition matrix \( A \), while also adding the process noise covariance \( Q \) to account for uncertainty introduced by the process noise \( w \).

\textbf{2) Correction}
\begin{align}
K_k &= P_{k|k-1} C^T (C P_{k|k-1} C^T + R)^{-1}, \label{eq:kalman_gain} \\
\hat{x}_{k|k} &= \hat{x}_{k|k-1} + K_k (z_k - C \hat{x}_{k|k-1}), 
\label{eq:state_update} \\
P_{k|k} &= (I - K_k C) P_{k|k-1}. \label{eq:covariance_update}
\end{align}
Here, \( I \) is the identity matrix, and \( R \) is the measurement noise covariance matrix aligning with the noise \( v \).  In \eqref{eq:kalman_gain}, the term \( C P_{k|k-1} C^T + R \) represents the total uncertainty in the measurement domain. Specifically, \( C P_{k|k-1} C^T \) is the predicted error covariance projected into the measurement space through the measurement matrix \( C \), while \( R \) accounts for the measurement noise covariance \cite{cordeiro_understanding_2021}. The Kalman gain \( K_k \) is then calculated as the ratio of the predicted covariance \( P_{k|k-1} C^T \) to this total uncertainty \cite{cordeiro_understanding_2021}. This construction balances the contributions of the model and the measurement, leading to a higher Kalman gain for smaller measurement uncertainty and equivalently to a higher weighting of the measurement. Equation~\eqref{eq:covariance_update} describes the update of the error covariance \( P_{k|k} \), which reflects the reduction in uncertainty after incorporating the measurement. 


By iteratively reducing the uncertainty in \( P_k \), the Kalman filter provides an optimal estimate of the state, assuming the system dynamics and noise characteristics are accurately modeled \cite{simon_optimal_2006}. If $Q$ is underestimated, the filter overweights the model and may fail to track changes accurately. Conversely, if $R$ is underestimated, the filter overweights noisy measurements. The error covariance matrix \( P \) in the Kalman filter converges to a steady-state value \( P_\infty \) when $Q$ and $R$ are in fact constant, and the Kalman gain \( K \) also stabilizes to \( K_\infty \). In the steady state, the filter operates optimally even without recalculating \( P \) and \( K \) at each step. \cite{welch_introduction_2006, simon_optimal_2006}

\subsection{Discrete Extended Kalman Filter}
The extended Kalman filter (EKF) extends the linear Kalman filter to handle systems where the dynamics and/or the measurement functions are nonlinear. It does so by approximating these functions through linearization around the current estimate, using partial derivatives of each \cite{welch_introduction_2006}. The system's state transition and measurement models are given by functions $f(\cdot)$ and $h(\cdot)$, here as per \eqref{eq:2ms_nonlin} and \eqref{eq:2ms_nonlinoutput}. 
As the system dynamics are described by nonlinear continuous equations, state propagation can be performed using a numerical ODE solver, allowing more accurate results through direct integration of the nonlinear dynamics. Alternatively, state propagation could be approximated using the linearized system, offering lower computational cost at the expense of accuracy \cite{pyrhonen_linearization-based_2023}. Error covariance propagation and calculation of the Kalman gain, regardless of the state propagation method, rely on the linearized system \cite{pyrhonen_linearization-based_2023}. Therefore, the Jacobians $A(t)$ and $C(t)$ are computed as the partial derivatives of $f(\cdot)$ and $h(\cdot)$ with respect to \mbox{the state \cite{simon_optimal_2006}:} 
\begin{equation}
A(t) = \left.\frac{\partial f}{\partial x}\right|_{\hat{x}(t),\,u(t)},
\quad \quad
C(t) = \left.\frac{\partial h}{\partial x}\right|_{\hat{x}(t)}.
\end{equation}
\vspace{-0.1cm}

For a discrete-time simulation, the Jacobian matrix $A$ can be discretized over the chosen time step $\Delta t$ using the exact discretization method $A_d = e^{A \Delta t}$ 
where the matrix exponential is computed numerically (e.g., using scaling and squaring or series expansion) \cite{el-kebir_discretization_2025,moler_nineteen_2003}. Thus, the discrete Kalman filter algorithm from Section~\ref{discreteKF} can be applied without further modifications.

\subsection{Disturbance Observer}
To use the Kalman filter framework as a DOB, the state vector is extended by introducing $x_5 = T_d$. As $T_d$ reflects the Kalman estimation of the unknown input $u_1$ according to equation \eqref{eq:2ms}, we can transform the matrices such that $u_1$ now affects $x_5$, aligning with \eqref{eq:pt1}, and $x_5$ affects $x_2$, respectively. Thus, the adapted system changes from \eqref{eq:2ms} to \eqref{eq:2ms_adapted}. Now, the estimation of $T_d$ can be used as input $u_1$, enabling the Kalman-based filter to estimate the driver torque modeled by \eqref{eq:pt1}. Whereas before, $u_1$ directly affected $x_2$, we now have the PT1 element in between, making the response to $u_1$ slightly slower, depending on $T$. Therefore, when simulatively generating the measurement values $z_1$ and $z_2$ with the known input $u$, the use of the initial model as per \eqref{eq:2ms} is more accurate, even though a fast time constant $T$ might be used and introduces a neglectable delay only.
\begin{equation}
\label{eq:2ms_adapted}
\setlength\arraycolsep{0.7pt}
\dot{x}
=
\begin{bmatrix}
0 & 1 & 0 & 0 & 0\\
\frac{-c_{g}}{J_{sw}} & \frac{-d_{g} - d_{sw}}{J_{sw}} & \frac{c_{g}}{J_{sw}} & \frac{d_{g}}{J_{sw}} & \frac{1}{J_{sw}} \\
0 & 0 & 0 & 1 & 0\\
\frac{c_{g}}{J_{m}} & \frac{d_{g}}{J_{m}} & \frac{-c_{g}}{J_{m}} & \frac{-d_{g} - d_{m}}{J_{m}} & 0 \\
0 & 0 & 0 & 0 & \frac{-1}{T} \\
\end{bmatrix}
x
+
\begin{bmatrix}
0 & 0 \\
0 & 0 \\
0 & 0 \\
0 & \frac{1}{J_{m}} \\
\frac{K}{T} & 0 \\
\end{bmatrix}
u
\end{equation}
\vspace{-0.1cm}

For the EKF-based DOB, the adaptation can be done analogously. The system of differential equations then changes from \eqref{eq:2ms_nonlin} to \eqref{eq:2ms_nonlin_adapted}, with the corresponding damping and stiffness terms~\eqref{eq:damping_terms}. 
\vspace{-0.1cm}
\begin{equation}
\label{eq:2ms_nonlin_adapted}
\begin{aligned}
\dot{x}_1 &= x_2 \\
\dot{x}_2 &= \frac{T_g^c + T_g^d - T_{sw}^d + x_5} 
{J_{sw}} \\
\dot{x}_3 &= x_4 \\
\dot{x}_4 &= \frac{-T_g^c -  T_g^d - T_{m}^d + u_2}{J_m} \\
\dot{x}_5 &= -\frac{1}{T} x_5 + \frac{K}{T} u_1 \\
\end{aligned}
\end{equation}
\vspace{-0.1cm}

This approach enables accurate estimation of driver torque, with a high-pass filter isolating the unintentional high-frequency components targeted by the disturbance observer.

\begin{figure*}[h!]
\centering
\includegraphics[width=0.98\textwidth]
{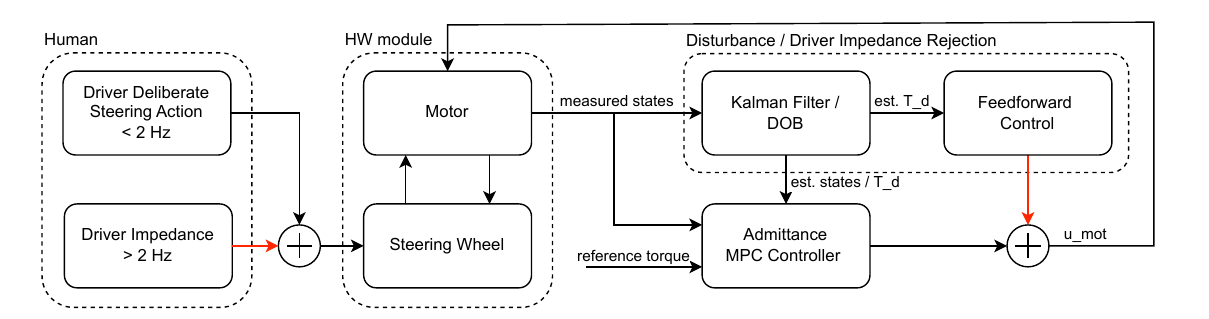}
\caption{Control structure of the HW module using MPC and Kalman filter-based DOB}
\label{fig:SbW1.drawio.pdf}
\end{figure*}

\section{Results}
\subsection{Simulation Setup}
The system architecture, comprising the DOB, the HW model and the model predictive controller (MPC), is schematically illustrated in Fig. \ref{fig:SbW1.drawio.pdf}, where the human input is separated by frequency and combined as a torque applied to the steering wheel. The mechanical coupling between the motor and the steering wheel, here modeled as a spring-damper, is indicated by the bidirectional arrows between the blocks. The driver impedance rejection block complements the admittance MPC, aiming to compensate for the high-frequency driver impedance indicated by the red arrows. 
The measurements $y_1$ and $y_2$ are generated using the initial \mbox{model \eqref{eq:2ms},} whereas the extended system \eqref{eq:2ms_adapted} is used within the filter.

The parameters used in the simulation are shown in \mbox{Table~\ref{tab:parameters}.} The HW module model parameters are primarily based on \cite{kazemi_comprehensive_2009} and \cite{ewald_regelung_2022}. With these parameters, the observability matrix of the linear system \eqref{eq:2ms_adapted} has full rank and the system is therefore observable. To simplify the simulation, the Stribeck friction curve is the sole nonlinear term included. As described in \ref{subsec:modeldriver}, the prediction for the driver torque $T_d$ only partially captures the driver's actual behavior. Consequently, a significantly higher process uncertainty is assigned to $T_d$. A comparable level of uncertainty is assumed for the other states.
\begin{table}[htbp]
\vspace{0.0cm}
\renewcommand{\arraystretch}{1.15}
\centering
\caption{Parameters: General, Linear, Nonlinear}
\label{tab:parameters}
\begin{tabular}{|c|c|c|}
\hline
\textbf{Parameter} & \textbf{Symbol} & \textbf{Value} \\ \hline
Sample time & $T_s$ & $0.001 \, \mathrm{s}$ \\ \hline
Sine freq. aktive drv trq & $f_{act}$ & $0.8 \, \mathrm{Hz}$ \\ \hline
Sine freq. passive drv trq & $f_{pas}$ & $7 \, \mathrm{Hz}$ \\ \hline
Inertia SW & $J_{sw}$ & $0.04 \, \mathrm{kg \cdot m^2}$ \\ \hline
Inertia motor & $J_{m}$ & $0.002 \, \mathrm{kg \cdot m^2}$ \\ \hline
PT1 time const. & $T$ & $0.08 \, \mathrm{s}$ \\ \hline
PT1 gain & $K$ & $1 \, \mathrm{}$ \\ \hline
IIR cutoff freq. & $F_{cut}$ & $4 \, \mathrm{Hz}$ \\ \hline
IIR order & $n$ & $1 \, \mathrm{}$ \\ \hline
Process Noise Cov. & $Q$ & $10^{-7}  \cdot \operatorname{diag}(1, 1, 1, 1, 10^{6})$ \\ \hline
Meas. Noise Cov. & $R$ & $10^{-6}  \cdot \operatorname{diag}(1, 1)$ \\ \hline
Friction coefficient gear & $d_g=d_g^1$ & $10^{-5} \, \mathrm{Nm/(rad/s)}$ \\ \hline
Spring coefficient gear & $c_g=c_g^1$ & $76.9731 \, \mathrm{Nm/rad}$ \\ \hline
\noalign{\hrule height 1.2pt}
Friction coefficient SW & $d_{sw}$ & $0.225 \, \mathrm{Nm/(rad/s)}$ \\ \hline
Friction coefficient motor & $d_m$ & $0.0034 \, \mathrm{Nm/(rad/s)}$ \\ \hline
\noalign{\hrule height 1.2pt}
Viscous fric. coef SW & $d_{sw}^v$ & $0.0084 \, \mathrm{Nm/(rad/s)}$ \\ \hline
Viscous fric. coef motor & $d_{m}^v$ & $0.0036 \, \mathrm{Nm/(rad/s)}$ \\ \hline
Static fric. torque SW & $d_{sw}^{s}$ & $0.735 \, \mathrm{Nm}$ \\ \hline
Static fric. torque motor & $d_{m}^{s}$ & $0.3150 \, \mathrm{Nm}$ \\ \hline
Kinetic fric. torque SW & $d_{sw}^{k}$ & $0.4620 \, \mathrm{Nm}$ \\ \hline
Kinetic fric. torque motor & $d_{m}^{k}$ & $0.1980 \, \mathrm{Nm}$ \\ \hline
Char angular velocity SW & $\omega_{sw}^{s}$ & $0.85 \, \mathrm{rad/s}$ \\ \hline
Char angular velocity motor & $\omega_{m}^{s}$ & $0.85 \, \mathrm{rad/s}$ \\ \hline
Delta Stribeck & $\delta$ & $2 $ \\ \hline
\end{tabular}
\vspace{0.0cm}
\end{table}

\subsection{Simulation Results}
\begin{figure}[H]
\centering
\includegraphics[width=0.48\textwidth]{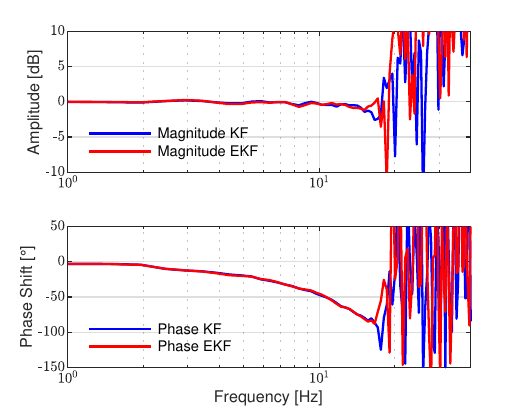}
\caption{Bode plots of the driver torque estimation for KF and EKF, identified via Matlab's \texttt{tfestimate}.}
\label{fig:bode_KF_EKF}
\end{figure}

\begin{figure}[htbp]
\centering
\includegraphics[width=0.48\textwidth]{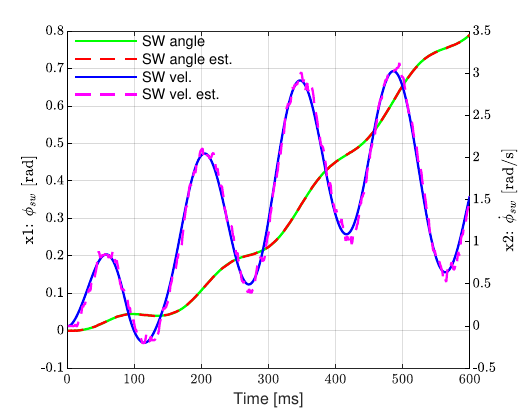}
\caption{Evolution of steering wheel angle and angular velocity of the nonlinear system and EKF-based estimations without disturbance rejection}
\label{fig:states_EKF_wo_rej.pdf}
\end{figure}

To assess the filter behavior, the transfer functions of the Kalman filter and Extended Kalman filter were identified via Matlab’s \texttt{tfestimate} using a chirp input. The resulting Bode plots, shown in Fig.~\ref{fig:bode_KF_EKF}, reveal the magnitude and phase response of the torque estimation. The phase lag increases with frequency, reaching approximately $35^\circ$ at $7\,\text{Hz}$. 
Beyond $15\,\text{Hz}$, the phase shift becomes excessive, limiting the estimation bandwidth in the present configuration. However, as road-induced feedback to the driver above $15\,\text{Hz}$ typically exhibits low amplitude, disturbance rejection is primarily required at lower frequencies, for which the available bandwidth is sufficient in practical steer-by-wire applications.
For a qualitative evaluation of the estimation performance, Fig.~\ref{fig:states_EKF_wo_rej.pdf} illustrates the evolution of the steering wheel states along with their EKF-based estimations, while Fig.~\ref{fig:torque_EKF_wo_rej.pdf} depicts the corresponding torques, including the driver torque estimation and its high-pass filtered signal. As shown, the estimated torque closely matches the actual torque, with an approximate delay of $14\,\text{ms}$, consistent with the $35^\circ$ phase lag observed at $7\,\text{Hz}$ in the Bode diagram. The motor torque is determined by a MPC that employs an impedance model to replicate the feel of conventional steering. Since disturbance rejection is inactive, high-frequency noise significantly impacts the steering wheel's motion, as visualized in Fig.~\ref{fig:torque_EKF_wo_rej.pdf}.

\begin{figure}[htbp]
\centering
\includegraphics[width=0.48\textwidth]{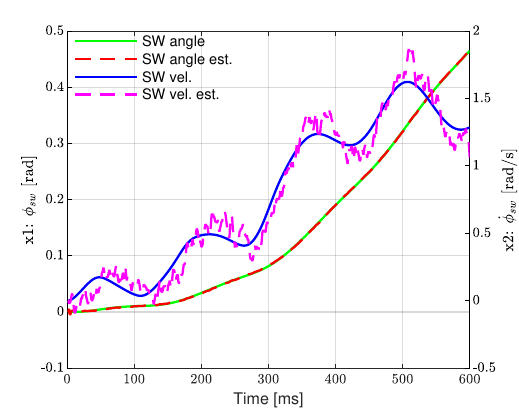}
\caption{Evolution of steering wheel angle and angular velocity of the nonlinear system and EKF-based estimations with disturbance rejection}
\label{fig:states_EKF_w_rej.pdf}
\end{figure}

Fig. \ref{fig:states_EKF_w_rej.pdf} and Fig. \ref{fig:torque_EKF_w_rej.pdf} illustrate the same scenario with EKF estimations, but with the DOB being incorporated into the MPC, resulting in effective disturbance rejection. Consequently, the motor torque counteracts the driver’s high-frequency noise, leading to a reduced impact on the steering wheel states. The steering wheel angle now predominantly follows the low-frequency, intentional steering maneuvers performed by the driver.

\begin{figure}[thbp]
\vspace{-0.0cm}
\centering
\includegraphics[width=0.48\textwidth]{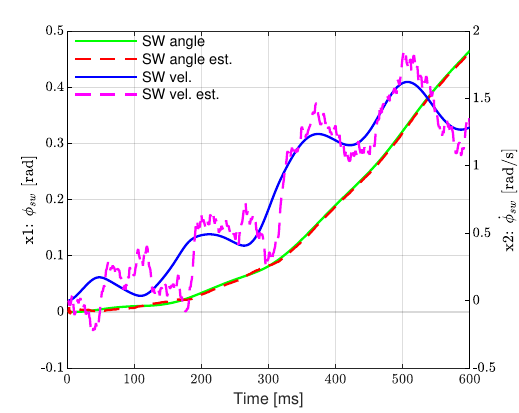}
\caption{Evolution of steering wheel angle and angular velocity of the nonlinear system and KF-based estimations with disturbance rejection}
\label{fig:states_KF_w_rej.pdf}
\end{figure}

\begin{figure}[thbp]
\centering
\includegraphics[width=0.48\textwidth]{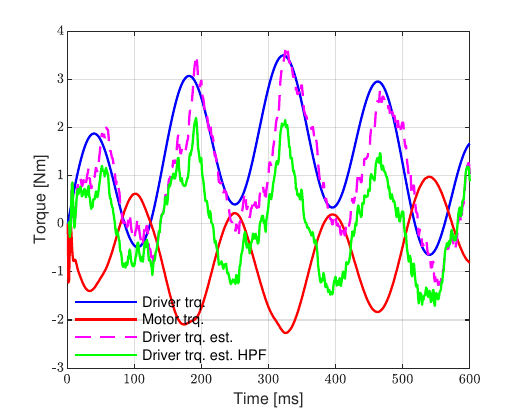}
\caption{Torque evolution of the nonlinear system and KF-based estimation of driver torque and its high pass filtered signal with disturbance rejection}
\label{fig:torque_KF_w_rej.pdf}
\end{figure}

To compare the quality of KF- and EKF-based estimations, the KF is applied to the nonlinear system. The results are shown in Fig. \ref{fig:states_KF_w_rej.pdf} and Fig. \ref{fig:torque_KF_w_rej.pdf}. Despite the significant nonlinearity, the KF provides reasonable state estimates, albeit with reduced accuracy. For angular velocity, the estimation quality of the KF degrades notably near zero, where nonlinear friction becomes relevant. This limitation is mitigated in the EKF due to its nonlinear prediction capability.
To complement the qualitative evaluation, the passive driver torque estimation performance can be quantified using the normalized root mean square error (RMSE) and mean absolute error (MAE), which reflect the average and worst-case deviation between estimated and true torque values. The EKF achieves lower errors than the KF when applied to the nonlinear system, with normalized RMSE and MAE of 11.96\,\% and 9.91\,\%, compared to 13.84\,\% and 11.16\,\% for the KF. This is consistent with the qualitative observations in Fig.~\ref{fig:torque_EKF_w_rej.pdf} and \ref{fig:torque_KF_w_rej.pdf}.

\section{Discussion}
The presented approaches for estimating the passive driver torque prove effective, even without using torque sensor information. The EKF provides a more accurate and dynamic estimation with small delay. Still, even with nonlinear dynamics of the HW module, the KF proves effective. Depending on the context, the advantages of the linear estimator might even outweigh the benefits of the nonlinear one. However, it has to be noted that the system's observability and, consequently, the quality of the estimation are highly dependent on the chosen parameters. The parameters used in this work result in a high 2-norm condition number for the observability matrix, indicating that the matrix is poorly conditioned and thus not well observable. While the setup performs well in simulation, the realism of the chosen parameters has not been thoroughly validated. As a result, it remains uncertain whether the approach will perform equally well in real-world scenarios, where factors such as noise, unmodeled dynamics, and other imperfections could further challenge observability.


\begin{figure}[htbp]
\centering
\includegraphics[width=0.48\textwidth]{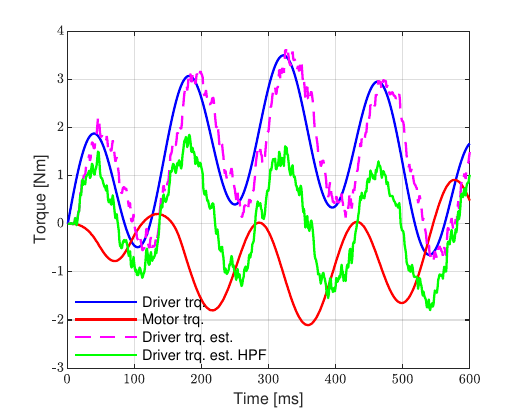}
\caption{Torque evolution of the nonlinear system and EKF-based estimation of driver torque and its high pass filtered signal without disturbance rejection}
\label{fig:torque_EKF_wo_rej.pdf}
\end{figure}
\begin{figure}[htbp]
\centering
\includegraphics[width=0.48\textwidth]{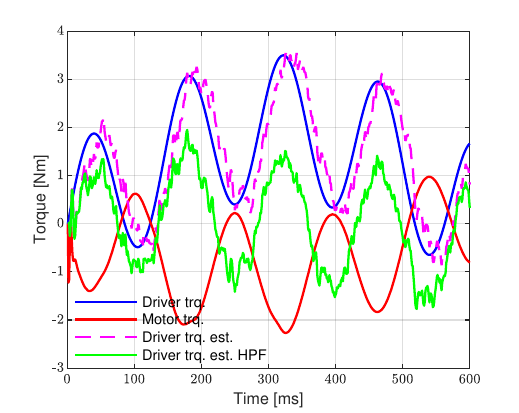}
\caption{Torque evolution of the nonlinear system and EKF-based estimation of driver torque and its high pass filtered signal with disturbance rejection}
\label{fig:torque_EKF_w_rej.pdf}
\end{figure}

Furthermore, it should be noted that, as shown in \mbox{Fig. \ref{fig:torque_EKF_wo_rej.pdf},} Fig. \ref{fig:torque_EKF_w_rej.pdf}, and Fig. \ref{fig:torque_KF_w_rej.pdf}, the IIR high-pass filter introduces a negative phase shift. This leads to an apparent non-causal compensation of the delay introduced by the Kalman-based estimation, which may artificially improve the results. However, with an observed delay of approximately $14 ~ \mathrm{ms}$, corresponding to $35^{\circ}$ phase lag at $7 ~ \mathrm{Hz}$, and remaining tuning potential, the approach demonstrates great performance detecting even high-frequency torques, suggesting that an effective use of such a signal is feasible.

\section{Conclusion and Outlook}
In this paper, a linear and nonlinear Kalman-based disturbance observer for driver induced high-frequency torque has been presented. For this, accurate linear and nonlinear models of the HW module have been derived, and a  PT1 element was used as driver torque model, enabling the Kalman-framework to estimate the driver torque. Both the KF and the EKF demonstrated good estimation performance, with EKF having advantages especially in nonlinear regions. Consequently, disturbance rejection could successfully be implemented in the SbW control, leading to the desired behavior of oversteering the driver impedance.

For future research, the robustness and stability of the observer need to be examined under more realistic operating conditions. Since the DOB relies heavily on the accuracy of the HW module model, it is critical to assess its sensitivity to parameter variations and modeling inaccuracies. In real-world environments, factors such as unmodeled dynamics, temperature-dependent parameter drift, and external disturbances can significantly impact estimation performance. Therefore, extensive parameter sensitivity studies and robustness analysis will be essential. To transition from simulation to industrial use, several practical challenges must be addressed. First, measurement and process noise levels encountered in physical systems are likely to differ from those assumed in simulation. The observer must maintain performance in the presence of sensor noise, quantization effects, and time-varying disturbances. Second, real-time operation introduces communication delays and network latency, particularly in distributed automotive architectures. These delays may degrade observer stability and should be accounted for in both design and implementation phases. Ultimately, a real-world testing roadmap will include: (1) hardware-in-the-loop (HiL) simulation with realistic sensor models and delays, (2) prototype implementation on automotive ECUs with profiling of execution time and memory usage, and (3) in-vehicle testing under a variety of driving scenarios and road conditions.

\newpage 
\bibliographystyle{IEEEtran}
\bibliography{sources.bib}

\end{document}